\documentclass[11pt,a4paper]{article}
\usepackage[hyperref]{emnlp2020}

\usepackage{color}

\usepackage{times}
\usepackage{latexsym}
\usepackage{amsmath}

\usepackage[russian,main=english]{babel}
\usepackage{graphicx}
\graphicspath{ {./figures/} }
\usepackage{enumitem}
\usepackage{booktabs}

\usepackage{microtype}

\aclfinalcopy 

\title{Zero-Shot Translation Quality Estimation with\\ Explicit Cross-Lingual Patterns}

\author{Lei Zhou$^\mathsection$~~~~Liang Ding$^{\dagger}$~~~~Koichi Takeda$^\mathsection$\\
  $^\mathsection$FVCRC, Graduate School of Informatics, Nagoya University\\
  {\tt \{zhou.lei@a.mbox,takedasu@i\}.nagoya-u.ac.jp} \\
  $^\dagger$UBTECH Sydney AI Centre, School of Computer Science\\ Faculty of Engineering,
  The University of Sydney\\
  {\tt ldin3097@uni.sydney.edu.au}}

\date{}

\begin{document}
\maketitle

\begin{abstract}
This paper describes our submission of the WMT 2020 Shared Task on Sentence Level Direct Assessment, Quality Estimation (QE). In this study, we empirically reveal the \textit{mismatching issue} when directly adopting BERTScore~\cite{Zhang:2019th} to QE. Specifically, there exist lots of mismatching errors between source sentence and translated candidate sentence with token pairwise similarity.
In response to this issue, we propose to expose explicit cross lingual patterns, \textit{e.g.} word alignments and generation score, to our proposed zero-shot models.
Experiments show that our proposed QE model with explicit cross-lingual patterns could alleviate the mismatching issue, thereby improving the performance. 
Encouragingly, our zero-shot QE method could achieve comparable performance with supervised QE method, and even outperforms the supervised counterpart on 2 out of 6 directions.
We expect our work could shed light on the zero-shot QE model improvement.
 
\end{abstract}

\section{Introduction}
Translation quality estimation (QE) ~\cite{blatz2004confidence,specia2018findings,specia2020findings} aims to predict the quality of translation hypothesis without golden-standard human references,  setting it apart from reference-based translation metrics. Existing reference-based evaluation metrics, e.g. BLEU~\cite{bleu}, METEOR~\cite{Banerjee:to}, NIST~\cite{on:wg}, ROUGE~\cite{out:tl}, TER~\cite{Snover:tz}, are commonly used in language generation tasks including translation, summarization, and captioning but all heavily rely on the quality of given references.

Recently, \citep{edunov2020evaluation} show that reference-based automatic evaluation metrics, e.g., BLEU, are not always reliable because the human translated references are translationese~\cite{koppel2011translationese,graham2019translationese}.
Thus, an automatic method with no access to any references, i.e., QE, is highly appreciated.

\begin{figure}
    \centering
    \includegraphics[width=0.45\textwidth]{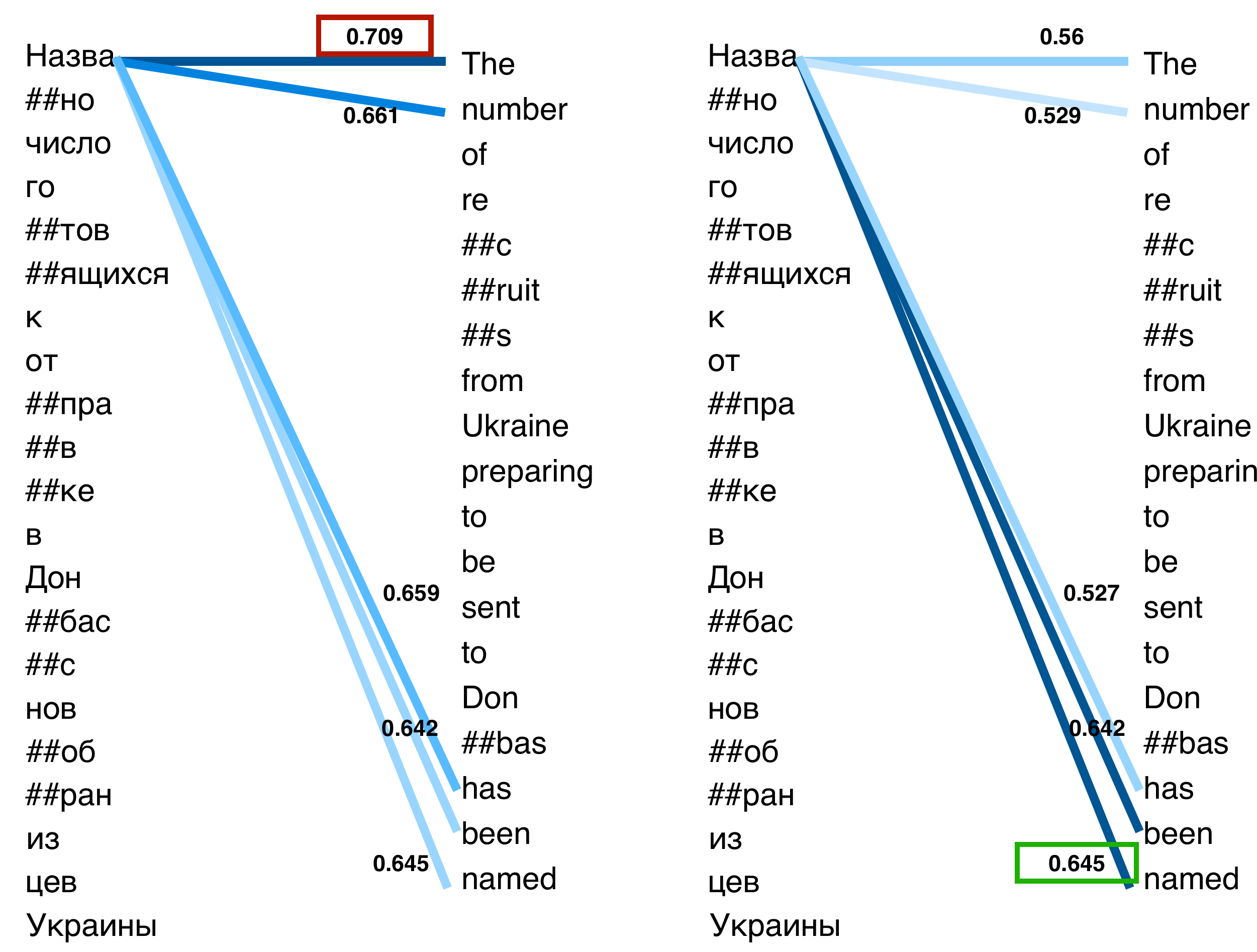}
    \caption{
    Example of mismatching error,  Russian$\rightarrow$English. On the left, token ``\begin{otherlanguage}{Russian}Назва\end{otherlanguage}'' is mismatched to ``The'' with the maximal probability (within the \textcolor{red}{red} rectangle) only. On the right, guided by our proposed cross-lingual patterns, ``\begin{otherlanguage}{Russian}Назва\end{otherlanguage}'' is correctly matched to the token ``named'' with the maximal probability (within the \textcolor{green}{green} rectangle.)}
    \label{fig:example-alignment}
\end{figure}

In this paper, we mainly focus on sentence level QE metrics, where existing studies categorize it into two classes: 1) supervised QE with human assessment as supervision signal: a feature extractor stacked with an estimator~\cite{Yankovskaya:2019wb,Wang:up,Fan:2019it}; 2) unsupervised QE without human assessment, which normally based on the pre-trained word embeddings, for example, YISI~\cite{Lo:2019uc} and BERTScore~\cite{Zhang:2019th}. Our work follows the latter, where we adopt BERTScore~\cite{Zhang:2019th} without extra fine-tuning. In particular, we implement our approach upon the pre-trained multilingual BERT~\cite{devlin2019bert} and XLM~\cite{Lample:2019tg}.

We first empirically reveal the \textit{mismatching issue} when directly adopting BERTScore~\cite{Zhang:2019th} to QE task. Specifically, there exist lots of mismatching errors between source tokens and translated candidate tokens when performing greedy matching with pairwise similarity. Figure~\ref{fig:example-alignment} shows an example of the mismatching error, where the Russian token ``\begin{otherlanguage}{Russian}Назва\end{otherlanguage}'' is mismatched to the English token ``The'' due to lacking of 
proper guidance. 

To alleviate this issue, we design two explicit cross-lingual patterns to augment the BERTScore as a QE metric:

\begin{itemize}[topsep=0pt,itemsep=0pt,parsep=0pt,partopsep=0pt]
\item \textsc{Cross-lingual alignment masking}: we design an alignment masking strategy to provide the  pairwise similarity matrix with extra guidance. The alignment is derived from GIZA++~\cite{Och:up}.

\item \textsc{Cross-lingual generation score}: we obtain the perplexity, dubbed \textit{ppl}, of each target side token by force decoding with a pre-trained cross-lingual model, e.g. multilingual BERT and XLM. This generation score is weighted added on the similarity score.
\end{itemize}

\section{Methods}

\begin{table*}

\centering
\begin{tabular}{clccccccc}
\toprule
\textbf{\#}& \textbf{Metrics} & \textbf{en-de} & \textbf{en-zh} & \textbf{ro-en} & \textbf{et-en} & \textbf{ne-en} & \textbf{si-en} & \textbf{ru-en}\\ 
\midrule
1&Baseline (test) & 0.146 & 0.190 & 0.685 & 0.477 & 0.386 & 0.374 & 0.548 \\
\midrule
2&BERT & \textbf{0.120} & 0.167 & 0.650 & 0.306 & 0.475 & - & \textbf{0.354} \\
3&BERT (align) & 0.091 & 0.170 & 0.672 & 0.307 & \textbf{0.478} & - & 0.340 \\
4&BERT (ppl) & 0.068 & 0.187 & 0.671 & 0.321 & 0.468 & - & 0.311 \\
5&BERT (align+ppl) & 0.099 & \textbf{0.189} & \textbf{0.677} & \textbf{0.324} & 0.477 & - & 0.332 \\
\bottomrule
\end{tabular}
\caption{\label{results_pearson} Pearson correlations with sentence-level Direct Assessment (DA) scores. The results of supervised baseline~\cite{kepler-etal-2019-openkiwi}, provided by the organizer, show it's agreement with DA scores on the test set of WMT20 QE. As DA scores on test set aren't available at this point, we report our experiment results on valid set. }
\end{table*}

\begin{table*}
\centering
\begin{tabular}{clccccccc}
\toprule
\textbf{\#}& \textbf{Metrics} & \textbf{en-de} & \textbf{en-zh} & \textbf{ro-en} & \textbf{et-en} & \textbf{ne-en} & \textbf{si-en} & \textbf{ru-en}\\ 
\midrule
1&BERT & \textbf{0.143} & 0.131 & 0.389 & 0.217 & 0.318 & - & \textbf{0.259} \\
2&BERT (align) & 0.122 & 0.133 & 0.422 & 0.219 & \textbf{0.322} & - & 0.251 \\
3&BERT (ppl) & 0.105 & 0.145 & 0.416 & 0.225 & 0.315 & - & 0.240 \\
4&BERT (align+ppl) & 0.132 & \textbf{0.152} & \textbf{0.439} & \textbf{0.228} & 0.320 & - & 0.247 \\
\bottomrule
\end{tabular}
\caption{\label{results_kendall} Kendall correlations with sentence-level Direct Assessment (DA) scores.}
\end{table*}

\subsection{BERTScore as Backbone}\label{bertscore}
A pre-trained multilingual model generates contextual embeddings of both source sentence and translated candidate sentence, such that this pair of sentences in different language can be mapped to the same continuous feature space. Given a source sentence $x=\langle x_1, \dots, x_k  \rangle$, 
 the model generates a sequence of vectors $\mathbf{\langle x_1, \dots, x_k \rangle}$ while the candidate $\hat{y} = \langle \hat{y}_1, \dots, \hat{y}_l \rangle$ is mapped to $\mathbf{\langle \hat{y}_1, \dots, \hat{y}_l\rangle}$. Different from the reference-based BERTScore, where they compute the pairwise similarity between reference sentence and translated candidate sentence,
 we calculate the pairwise similarity between the source and translated candidate with dot-production, i.e., $\mathbf{x_i^\top\hat{y}_j}$.  We adopt greedy matching to force each source token to be matched to the most similar target token in the translated candidate sentence. The QE function based on BERTScore backbone therefore can be formulated as:
 
 \begin{equation}
\begin{aligned}
R_{\mathrm{BERT}} =\frac{1}{|x|} \sum_{x_i \in x}   \max_{\hat{y}_j \in \hat{y}} \mathbf{x_i^\top\hat{y}_j}, \\
P_{\mathrm{BERT}} = \frac{1}{|\hat{y}|}  \sum_{\hat{y}_j\in \hat{y}}   \max_{x_i\in x}  \mathbf{x_i^\top\hat{y}_j}, \\
F_{\mathrm{BERT}} = 2\frac{P_{\mathrm{BERT}} \cdot R_{\mathrm{BERT}}}{P_{\mathrm{BERT}} + R_{\mathrm{BERT}}}.
\end{aligned}
\end{equation}
where $R_{\mathrm{BERT}}$, $P_{\mathrm{BERT}}$ and $F_{\mathrm{BERT}}$ are inherited from \citet{Zhang:2019th}, representing Recall rate, Precision rate and F-score, respectively.

\subsection{Alignment Masking Strategy}
With aforementioned QE function, we can follow \citet{Zhang:2019th} to obtain the distance between the source sentence and translated candidate sentence via directly adding up the maximum similarity score of each token pair. However, because there exist lots of mismatching errors (as shown in Figure \ref{fig:example-alignment}), above sentence-level similarity calculation may be sub-optimal.
Moreover, \newcite{Zhang:2019th}'s calculation is suitable for mono-lingual scenario, which may be insensitive for cross-lingual computation. 
Thus, we propose to augment our QE metric with more cross-lingual signals. 

Inspired by~\citet{ding-etal-2020-self}, where they show it's possible to augment cross-lingual modeling by leveraging cross-lingual explicit knowledge.
we therefore employ word alignment knowledge from external models, e.g., GIZA++\footnote{\url{https://github.com/moses-smt/giza-pp}}, as additional information.

\begin{figure}
    \centering
    \includegraphics[width=0.45\textwidth]{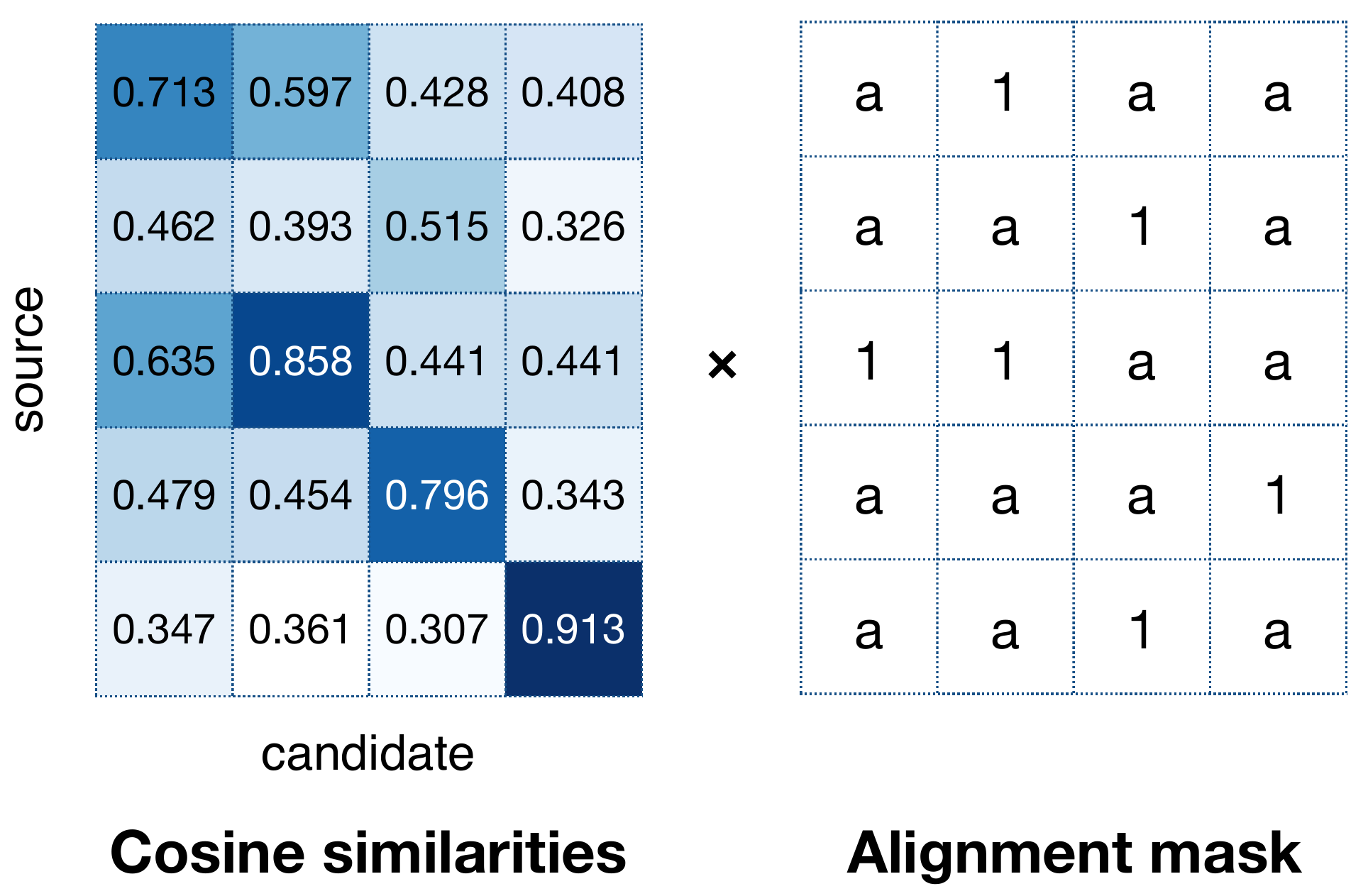}
    \caption{Word alignment as a mask matrix}
    \label{alignment_mask}
\end{figure}

\paragraph{Alignment masking} Both BERT~\cite{devlin2019bert} and XLM~\cite{Lample:2019tg} utilize BPE tokenization~\cite{Sennrich:2015uma}. It should be noted that in this paper, by word alignment we mean alignment of BPE tokenized word and subword units. Given a tokenized source sentence $x$ and candidate sentence $\hat{y}$,  alignment~\cite{Och:up} is defined as a subset of the Cartesian product of position,  $\mathcal{A} \subseteq \{(i,j): i=1,\dots,k;j=1,\dots,l \}$. Alignment results represented by $\mathcal{M}$ is defined as:
\begin{equation}
\mathcal{M} = \left\{ 
\begin{array}{cl} 
1 & (i,j) \in \mathcal{A} \\
0 \leq a \le 1 & \mathrm{otherwise}
\end{array}\right.
\label{eq:penalty}
\end{equation}
$\mathcal{M}$ is a penalty function over the similarity of unaligned tokens. It's a mask like matrix to assign a penalty weight $a$ \footnote{In our preliminary studies, $a=0.8$ picking from $\{0.0,0.2,0.4,0.8,1.0\}$ performs best, which then leaves as the default setting in the following experiments.} to the similarity of unaligned tokens while keeping that of aligned ones unchanged, as illustrated in Figure \ref{alignment_mask}. Thus, greedy matching is performed on a renewed similarity matrix, which is defined as the average of $\mathbf{x_i^\top\hat{y}_j}$ and masked $\mathbf{x_i^\top\hat{y}_j}$ by word alignment. For example, $R_{\mathrm{BERT}}$ is changed into:
\begin{equation}
R_{\mathrm{BERT(align)}} =\frac{1}{2|x|} \sum_{x_i \in x}   \max_{\hat{y}_j \in \hat{y}} (\mathbf{x_i^\top\hat{y}_j} +   \mathcal{M} \cdot \mathbf{x_i^\top\hat{y}_j} )
\label{eq:align}
\end{equation}
which can be characterized as balancing our proposed extra explicit cross-lingual patterns, i.e., word alignment. 

\subsection{Generation Score}
In additional to token similarity score, we introduce force-decoding perplexity of each target token as a cross-lingual generation score. For better coordination and considering our cross-lingual setting, we use the same pre-trained cross-lingual model, e.g. multilingual BERT, for both token embedding extraction and masked language model (MLM) perplexity generation. This cross-lingual generation score is added as:
\begin{equation}
F_{\mathrm{BERT(ppl)}} = (1 - \lambda)*F_{\mathrm{BERT}} + {\lambda}*ppl_{\mathrm{MLM}}
\label{eq:ppl}
\end{equation}
where the $\lambda$ can be seen as a variable that regulates the interpolation ratio between $F_{\mathrm{BERT}}$ and our proposed $ppl_{\mathrm{MLM}}$, making the generation score after combination more wisely. The effect of $\lambda$ will be discussed in the experiments.

\section{Experimental Results}
\subsection{Data}
Main experiments were conducted on the WMT20 QE Shared Task, Sentence-level Direct Assessment language pairs. The task contains 7 directions, including:
\begin{itemize}
    \itemsep=1pt
    \item English$\rightarrow$German (\textbf{en-de})
    \item English$\rightarrow$Chinese (\textbf{en-zh})
    \item Romanian$\rightarrow$English (\textbf{ro-en})
    \item Estonian$\rightarrow$English (\textbf{et-en})
    \item Nepalese$\rightarrow$English (\textbf{ne-en})
    \item Sinhala$\rightarrow$English (\textbf{si-en})
    \item Russian$\rightarrow$English (\textbf{ru-en})
\end{itemize} Each of them consists of 7K training data, 1K validation data and 1K test data. 

\subsection{Setup}
Based on our proposed QE metric in Section~\ref{bertscore}, we conduct the validataion and main experiments with two pre-trained cross-lingual model: bert-base-multilingual-cased\footnote{\url{https://huggingface.co/bert-base-multilingual-cased}} (12-layer, 768-hidden, 12-heads, trained on 104 languages) and xlm-mlm-100-1280\footnote{\url{https://huggingface.co/xlm-mlm-100-1280}} (16-layer, 1280-hidden, 16-heads, trained on 100 languages) for both contextual embedding representation and generation score. The 9th layer of multilingual BERT and the 11th of XLM are used to generate contextual embedding representations. 
Furthermore, we obtain bidirectional word alignment of all the training, validation and test dataset with GIZA++.
Notably, this work is a zero-shot approach that doesn't involve training on Direct Assessment (DA) scores, which makes our method suitable for real industry scenarios.

\begin{figure}
    \centering
    \includegraphics[width=0.46\textwidth]{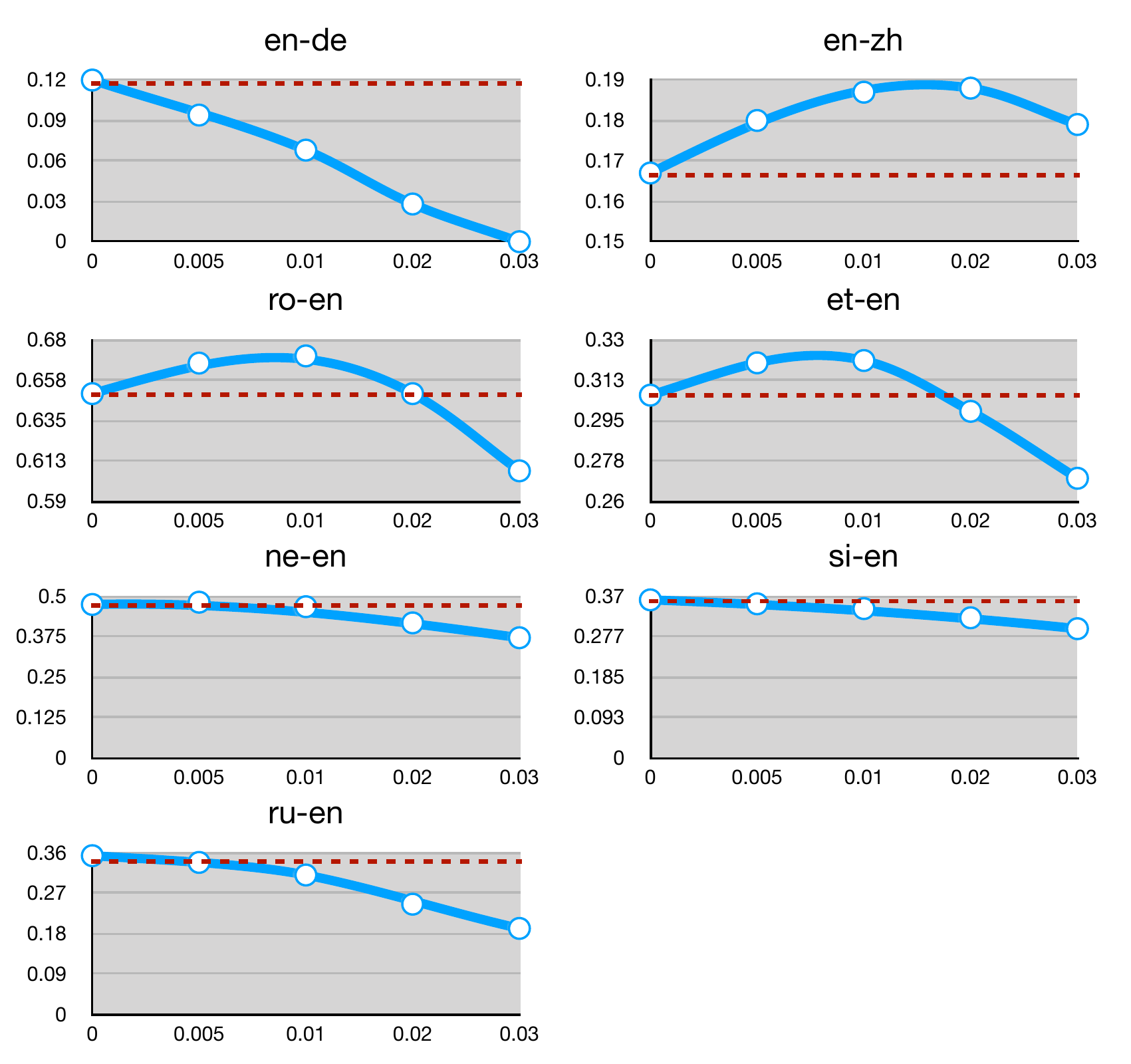}
    \caption{Pearson correlations with Direct Assessment (DA) scores when $\lambda \in \mathrm{[0,0.03]}$.}
    \label{lambda_of_ppl}
\end{figure}
\subsection{Ablation Study}\label{ablation}
In order to maximize the advantages of our proposed method for zero-shot translation QE, we conducted extensive ablation studies. We report the results of ablation studies on the validation dataset.

\paragraph{Effect of $\lambda$}
We conduct ablation studies to empirically decide the value of of $\lambda$ in Equation \ref{eq:ppl} when introducing generation scores. We observe positive effect of proper weighted additional generation score on \textbf{en-zh}, \textbf{ro-en}, \textbf{et-en}, \textbf{ne-en}, \textbf{si-en}. As illustrated in Figure ~\ref{lambda_of_ppl}, considering the average performance, we pick $\lambda=0.01$ from $[0,0.03]$.

\paragraph{Effect of different pretrained models}
We also investigated the effect to deploy our proposed fixed cross-lingual patterns on different state-of-the-art large scale pre-trained models, e.g., XLM~\cite{Lample:2019tg} (xlm-mlm-100-1280), BERT~\cite{Zhang:2019th} (bert-base-multilingual-cased). Table~\ref{bert_vs_xlm} lists a comparison of multilingual BERT and XLM in terms of the Pearson correlations with Direct Assessment (DA) scores. As seen, multilingual BERT outperforms XLM on almost all language pairs, excepting for \textbf{si-en}. One possible reason is that multilingual BERT is not pre-trained on Sinhala corpus while XLM does. In this end, we generate our final submission with XLM in \textbf{si-en} direction, and with multilingual BERT in other directions.

\begin{table}
\centering
\begin{tabular}{ccc}
\toprule
 & mBERT & XLM \\
\midrule
\textbf{en-de} & 0.120 & 0.056 \\
\textbf{en-zh} & 0.167 & 0.008 \\
\textbf{ro-en} & 0.650 & 0.568 \\
\textbf{et-en} & 0.306 & 0.254 \\
\textbf{ne-en} & 0.475 & 0.398 \\
\textbf{si-en} & - & 0.362 \\
\textbf{ru-en} & 0.354 & 0.228 \\
\bottomrule
\end{tabular}
\caption{\label{bert_vs_xlm} This is a comparison between multilingual BERT (``mBERT'') and XLM in terms of the Pearson correlations with Direct Assessment (DA) scores. Multilingual BERT performs better than XLM.}
\end{table}

\subsection{Main Results}

In the main experiments, we evaluate the agreement of our approach with Direct Assessment (DA) scores on validation dataset, as DA scores of the test set are not available at this point. Baseline results, which are evaluated on test set though, are also listed for general comparison.

As shown in Table~\ref{results_pearson}, our method could achieve improvements on 4 out of 6 directions, including \textbf{en-zh}, \textbf{ro-en}, \textbf{et-en} and \textbf{ne-en}. 
Particularly, combination of two strategies, i.e., \textsc{cross-lingual alignment} and \textsc{cross-lingual generation score}, could achieve better performance on \textbf{en-zh}, \textbf{ro-en} and \textbf{et-en} directions. 

Besides Pearson correlations, we also calculated Kendall correlations for all language pairs. As seen in Table~\ref{results_kendall}, the trends of Kendall correlations are same as Pearson correlations, validating the effectiveness of our proposed methods.

\subsection{Official Evaluations}
The official automatic evaluation results of our sub-missions for WMT 2020 are presented in Table~\ref{submission}. We participated QE (Sentence-Level Direct Assessment) in following language pairs: \textbf{en-de}, \textbf{en-zh}, \textbf{ro-en}, \textbf{ne-en}, \textbf{si-en}, \textbf{ru-en}, except for \textbf{et-en}. 
From the official evaluation results~\cite{specia2020findings} in terms of absolute Pearson Correlation, our submission achieves higher performance than supervised baseline~\cite{kepler-etal-2019-openkiwi} in \textbf{ne-en} and \textbf{si-en} (As shown in Table~\ref{submission}). 

Encouragingly, our proposed zero-shot QE metric could achieve comparable performance with supervised QE method, and even outperforms the supervised counterpart on 2 out of 6 directions.

\begin{table}
\centering
\begin{tabular}{ccc}
\toprule
 & Ours & \newcite{kepler-etal-2019-openkiwi} \\
\midrule
\textbf{en-de} & 0.111 & 0.146 \\
\textbf{en-zh} & 0.085 & 0.190 \\
\textbf{ro-en} & 0.650 & 0.685 \\
\textbf{et-en} & - & - \\
\textbf{ne-en} & \textbf{0.488} & 0.386 \\
\textbf{si-en} & \textbf{0.388} & 0.374 \\
\textbf{ru-en} & 0.400 & 0.548 \\
\bottomrule
\end{tabular}
\caption{\label{submission} Comparison of our submission and supervised baseline~\cite{kepler-etal-2019-openkiwi} on WMT20 sentence-level QE official test set, in terms of Pearson correlations.}
\end{table}

\section{Related Work}
\paragraph{MT evaluation}
Taking sentence-level evaluation as an example, reference-based metrics describe to which extend a candidate sentence is similar to a reference one~\cite{Sellam:2020tr}. BLEU~\cite{bleu}, METEOR~\cite{Banerjee:to}, NIST~\cite{on:wg}, ROUGE~\cite{out:tl} measure such similarity through n-gram matching, which is restricted to the exact form of sentences. TER~\cite{Snover:tz} and CHARACTER~\cite{Wang:up} use edit distance at word or character level to indicate the distance between candidate and reference. Different from these  
metrics that are restricted to the exact form of sentences, recent dominated neural model metrics
\textit{learn} to evaluate with human assessment as supervision signal, such as BEER~\cite{Stanojevic:wk} and RUSE~\cite{Shimanaka:usa}, or others as YiSi~\cite{Lo:2019uc} and BERTScore~\cite{Zhang:2019th} , evaluate with pre-trained word embedding, without using human assessment.
\paragraph{Incorporating Explicit Knowledge}
Several approaches have incorporated pre-defined or learned features into neural networks.
\newcite{Tai2015ImprovedSR} demonstrate that incorporating structured semantic information could enhance the representations. \newcite{sennrich-haddow:2016:WMT} feed the encoder cell combined embeddings of linguistic features including lemmas, subword tags, etc. 
\newcite{ding2017combining} leverage the domain knowledge to perform data selection to improve the machine translation models. \newcite{ding2019recurrent} incorporate the structure patterns of sentences, i.e., syntax, into the Transformer network to enhance seq2seq modeling performance. \newcite{aless2020fixed} utilize the pre-defined fixed patterns to replace the attention weights and show promising results. Inspired by above works, we propose to augment zero-shot QE model with cross-lingual patterns.

\section{Conclusion and Future Work}
In this work, we revealed a mismatching issue in zero-shot QE modeling. To alleviate it, we introduced two explicit cross-lingual patterns based on BERTScore backbone. Extensive experiments indicated that our proposed patterns, without fine-tuning, the QE model can be improved marginally. Notably, our zero-shot QE method outperforms supervised QE model on 2 out of 6 directions, shedding light on zero-shot QE researches.

In the future, we plan to explore more strategies for incorporating various auxiliary information and better in-domain fine-tuning~\cite{gururangan2020don} or introduce an non-autoregressive refiner~\cite{Wu2020SlotRefineAF} to address our revealed \textit{mismatching issue}. Also, it will be interesting to apply QE metrics on document-level machine translations with considering the dropped pronoun~\cite{wang2016novel,wang2018translating}.

\section*{Acknowledgments}
The authors wish to thank the anonymous reviewers for their insightful comments and suggestions.  

\bibliographystyle{acl_natbib}
\bibliography{emnlp2020}
\end{document}